\documentclass[fleqn,10pt]{wlscirep}
\usepackage[utf8]{inputenc}
\usepackage{marvosym}
\usepackage{multirow}
\usepackage[T1]{fontenc}

\usepackage{ragged2e}
\usepackage{lipsum}

\title{A Two-Stage Multi-Scale Attention-Based Network for Weakly Supervised Cataract Fundus Image Enhancement}

\author[1]{Xiaoyong Fang}
\author[2]{Yue Wang}
\author[2]{Xiangyu Li}
\author[2,*]{Wanshu Fan}
\author[2,3,*]{Dongsheng Zhou}

\affil[1]{Department, School of Safety and Management Engineering, Hunan
Institute of Technology, Hengyang, 421002, China}
\affil[2]{National and Local Joint Engineering Laboratory of Computer Aided Design, School of Software Engineering, Dalian, 116622, China. 
}
\affil[3]{School of Computer Science and Technology, Dalian
University of Technology, Dalian, 116024, China.}
\affil[*]{corresponding. fanwanshu@dlu.edu.cn, zhouds@dlu.edu.cn}


\begin{abstract}
Cataract is a major cause of vision loss and hinders further diagnosis. However, cataract fundus image enhancement often grapples with challenges such as limited paired cataract retinal images and insufficient recovery of fine details in the retinal images. To mitigate these challenges, we in this paper propose a two-stage multi-scale attention-based network (TSMSA-Net) for weakly supervised cataract fundus image enhancement. Our TSMSA-Net leverages the domain transformation to synthesis paired real-like cataract images, solving the problem of difficult acquisition of paired images. To further extract detailed information from fundus images and reduce the generation of artifacts during the enhancement process, we propose a multi-scale attention-based stage to learn more useful features for cataract image enhancement. Experimental results on Kaggle and ODIR-5K demonstrate that our TSMSA-Net outperforms current state-of-the-art cataract fundus images enhancement even without paired images and exhibits certain generalization ability.
  Experimental results on Kaggle and ODIR-5K datasets indicate that our TSMSA-Net outperforms the current state-of-the-art methods for cataract fundus image enhancement, even in the absence of paired images. Additionally, it demonstrates a certain level of generalization capability. The enhancement also can improve the performance of vessel segmentation and classification in cataract images.
\end{abstract}
\begin{document}

\flushbottom
\maketitle

\section*{Introduction}\label{sec1}
\begin{figure*}[!t] % Two column figure (notice the starred environment)
\centerline{\includegraphics[width=0.9\linewidth]{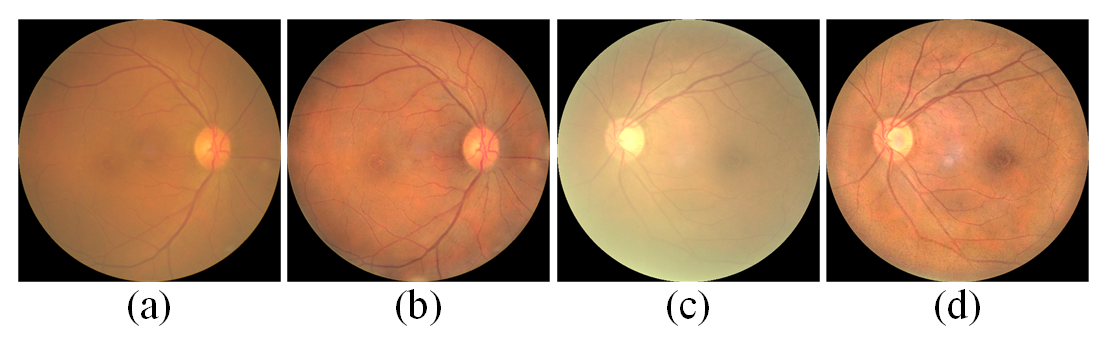}}
\caption{Fundus images. (a) and (b) are cataract images. (c) and (d) are the image enhanced by our TSMSA-Net. }
\label{fig:p6}
\end{figure*}

With the development of deep learning techniques, researchers have proposed numerous retinal disease detection and segmentation algorithms to aid in clinical diagnosis \cite{ozbay2023active}. However, these algorithms require high-quality retinal image inputs, while cataract retinal images are typically characterized by capture device and patient variability, making it difficult to ensure image quality. The quality of cataract retinal images is often shown as blurriness and poor image readability, Rendering the diagnosis of diseases by ophthalmologists challenging. Meanwhile, these poor-quality retinal images also may lead to suboptimal outcomes in automatic image processing, such as disease detection and segmentation, consequently impacting further disease diagnosis. Therefore, the restoration of cataract fundus images has clinical value \cite{peli1987enhancement}. Figure \ref{fig:p6} shows the fundus images of cataract image and the enhancement image restored by our TSMSA-Net. It can be observed that compared to Figure \ref{fig:p6}(b) and Figure \ref{fig:p6}(d), Figure \ref{fig:p6}(a) and Figure \ref{fig:p6}(c) are relatively blurry, with lower visibility of the fundus structures. It is difficult to accurately extract the fundus information of these blurred images, indicating the effectiveness of our TSMSA-Net. 

In addressing the blurriness of the cataract retinal images, researchers have extensively explored retinal image enhancement methods \cite{alwazzan2021hybrid,dash2022guidance,li2021applications,wangmo1,wangmo2}, utilizing classical methods to improve image quality. However, these manually designed algorithms fail to adequately preserve image details and suffer from the issue of amplifying image noise, leading to erroneous guidance in image restoration. Subsequently, with the advancement of deep learning, researchers began to utilize deep learning to enhance retinal images, achieving promising results \cite{wu2023fundus,li2022annotation,li2023generic,wangmo3}. 

Although the artificial synthesis methods \cite{liu2022degradation} can effectively obtain a large number of paired cataract retinal images, the synthetic function cannot fully cover the degradation conditions of cataract images, resulting in poor generalization ability of well-trained networks in real cataract retinal image application scenarios. In addition, unsupervised methods \cite{li2022annotation,yang2023retinal} are prone to losing detailed information of retinal images during the enhancement process due to the lack of supervised constraints.

To address the above challenges, we propose a two-stage multi-scale attention-based weakly supervised  cataract retinal image enhancement network (TSMSA-Net). Our TSMSA-Net mainly leverages a real-like cataract image synthesis stage to simulate more realistic paired degradation images. Specifically, we firstly apply a synthetic function to narrow the domain gap between synthetic and real cataract retinal images. Subsequently, we use the synthetic cataract images as the source domain and real cataract images as the target domain for domain transformation, resulting in real-like paired synthetic cataract retinal images. To further enhance the detail rendition, we introduce a multi-scale attention-based cataract retinal image enhancement stage. Unlike the widely used U-Net \cite{ronneberger2015u}, we utilize multi-scale attention modules to extract more abundant image detail features, avoiding the loss of details during the down-sampling process.

We summarise the main contributions of this paper as follows:
%\vspace{-3mm}
\begin{itemize}
    \item We propose a real-like cataract image synthesis stage to obtain more paired realistic synthesized cataract images, addressing the problem of difficult acquisition of paired images and the inability of the synthesis function to effectively cover the degradation of cataract images.

    \item We propose a multi-scale attention-based cataract image enhancement stage to better fuse multi-scale features, enabling the enhancement network to better recover image details and reduce the generation of artifacts.

    \item Qualitative and quantitative experiments on the Kaggle and ODIR-5K datasets demonstrate that our TSMSA-Net outperforms existing state-of-the-art cataract image enhancement methods. And the enhancement can improve the automatic image processing, such as disease classification and segmentation.
\end{itemize}

\section*{Related Work}\label{sec2}
\subsection*{Classical Retinal Image Enhancement Methods}
Classical methods for retinal image enhancement typically involve manually designing algorithms using prior information of the images, often focusing on enhancing contrast and brightness. For instance, many methods utilize contrast limited adaptive histogram equalization (CLAHE) to improve image contrast and achieve retinal image enhancement \cite{pizer1987adaptive}. Additionally, some researchers employ filtering techniques for image enhancement \cite{zhou2017color}. In addition, Dash et al. \cite{dash2022guidance} propose a joint model of fast guided filter and matching filter to enhance vascular extraction performance. Mohammed et al. \cite{alwazzan2021hybrid} present a hybrid algorithm that utilizes wiener filtering and CLAHE to enhance color retinal fundus images, reducing noise generation and achieving better enhancement effects.

Although the aforementioned methods can achieve excellent enhancement results, they also exist some limitations. Firstly, they are unable to precisely control the enhancement level, resulting in extracted features that struggle to preserve image details and are prone to amplifying image noise. Secondly, these manually designed prior information is often simplistic and cannot fully adapt to the various retinal degradations present in the real world, thus limiting their applicability.

\subsection*{Deep Learning-Based Retinal Image Enhancement Methods}
In recent years, deep learning has shown outstanding performance and has been widely applied to low-level visual tasks, such as segmentation \cite{fenge1}, dehazing \cite{quwu1}, and image enhancement \cite{diguang1}. In the field of image enhancement, there are also many methods that have achieved excellent enhancement effects \cite{peng2023u,gao2023ctcnet,jiang2021enlightengan}. However, most methods use the supervised learning method, leveraging a large amount of paired training data to learn the mapping from low-quality images to high-quality images \cite{deng2022rformer}. However, obtaining the paired data in medical scenarios is extremely difficult and time-consuming. Therefore, researchers have proposed artificially degrading high-quality images to synthesize low-quality images for supervised training \cite{DBLP:journals/bspc/RajST22,shen2020modeling,luo2020dehaze,li2023generic,liu2022degradation}. 

Since the degradation conditions that can be covered by artificially degraded methods are limited and cannot fully restore the degradation of real cataract images, and there exists a domain gap between artificially synthesized fundus degradation images and real degraded fundus images, the models trained by these methods often lack generalization ability when processing real degraded fundus images. Subsequently, some researchers propose semi-supervised methods for fundus image enhancement to reduce the dependence on the requirement of paired data. Wu et al. \cite{wu2023fundus} propose a semi-supervised generative adversarial network (SSGAN-ASP) to train the network using both supervised data and unsupervised data.
In addition, there are also methods proposed to use unpaired images for unsupervised learning in fundus image enhancement to reduce the need for paired data. Yang et al. \cite{yang2023retinal} introduce an unpaired fundus image enhancement method based on high-frequency extraction and feature description to preserve the structural information of the image and reduce the generation of vascular-like artifacts during the enhancement process. Li et al. \cite{li2022annotation} propose an unsupervised cataract fundus image restoration network (ArcNet) that does not require annotations. However, due to the lack of supervision constraints, unsupervised learning methods mainly simulate the results of high-quality images from low-quality images through image style transformation, which easily leads to the loss of detailed information in low-quality images.

\section*{Proposed Approach}\label{sec3}

\begin{figure*}[t] % Two column figure (notice the starred environment)
\includegraphics[width=1\linewidth]{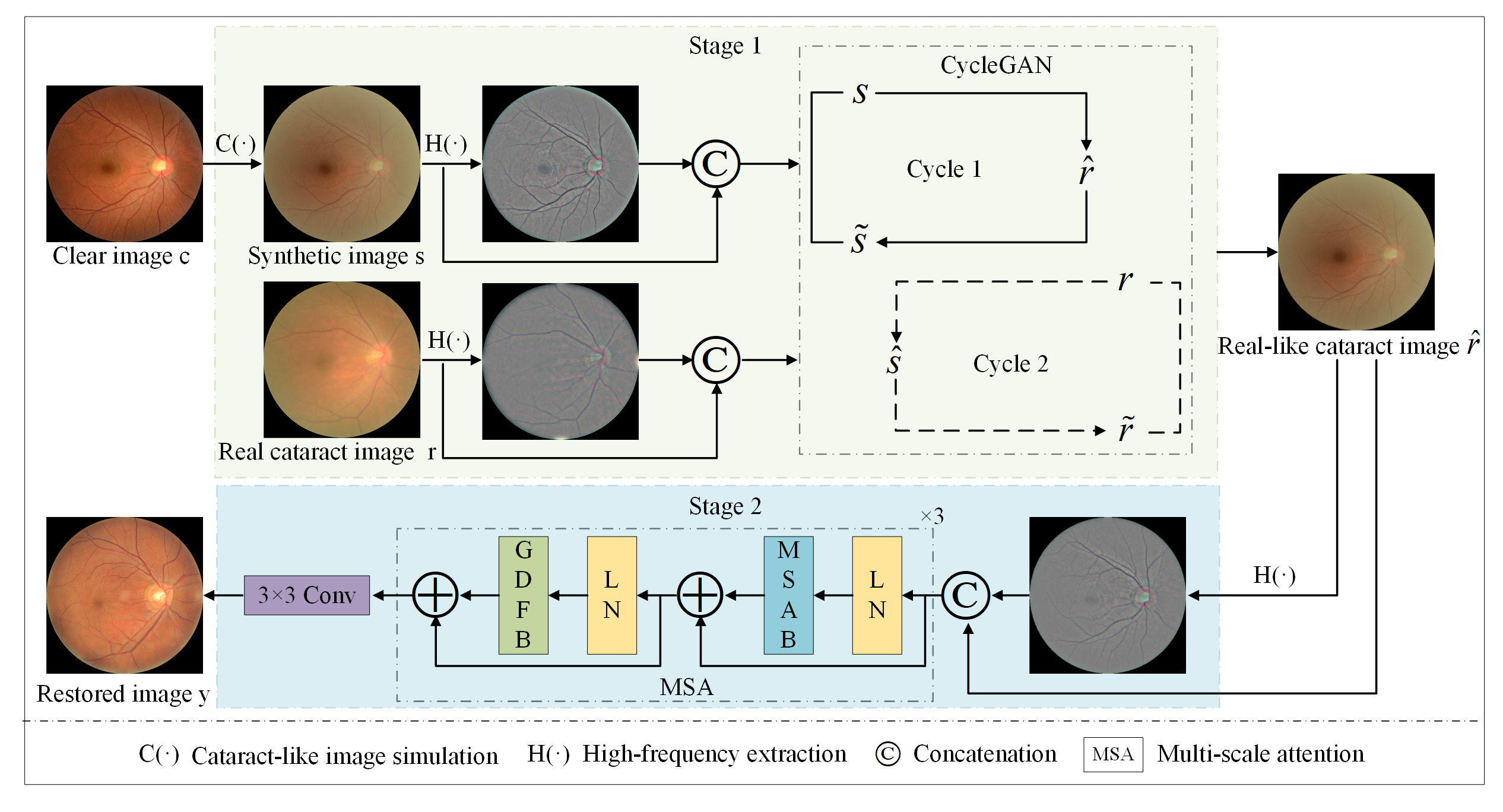}
\caption{The overall architecture of the proposed network. Our network consists of stage 1 and stage 2. In  stage 1, given a clear image c and a real cataract image r, we first extract structural information from each image for guidance. Subsequently, we employ CycleGAN \cite{zhu2017unpaired} to perform domain transformation on the synthetic image and the real cataract image, resulting in the generation of real-like cataract image. And in stage 2, we use a series of multi-scale attention (MSA) to extract more details from the synthetic real-like cataract images.}
\label{fig:p1}
\end{figure*}
Due to the inability of synthesized images generated by the composite function to fully simulate the degradation of real cataract fundus images, and the risk of losing structural and detailed information in unsupervised learning methods, we propose a two-stage multi-scale attention-based weakly supervised cataract fundus image enhancement network (TSMSA-Net). 
To better obtain synthesized cataract fundus images that are closer to the realistic scenarios, we propose a real-like cataract fundus image synthesis stage, introduced.
To better capture the detailed information of cataract fundus images, we introduce a multi-scale attention-based stage to learn the features of cataract fundus images under supervised guidance, aiming for better enhancement of the details in cataract fundus images, detailed.

\subsection*{Overall Pipeline}
Figure \ref{fig:p1} illustrates the overall architecture of our two-stage multi-scale attention-based weakly supervised cataract retinal image enhancement network (TSMSA-Net), which consists of a real-like cataract fundus image synthesis stage as stage 1 and a multi-scale attention-based cataract fundus image enhancement stage as stage 2. 
In stage 1, we first utilize a composite function C\(\left( \cdot \right)\) to generate simulated cataract fundus images, aiming to reduce the domain gap between the synthesized and real cataract fundus images. Subsequently, we use the CycleGAN \cite{zhu2017unpaired} network for domain translation between the synthesized cataract fundus images and the corresponding real cataract fundus images, to make the synthesis images are closer to the real ones. We utilize these synthesized cataract fundus images paired with their corresponding clear fundus images as the input for the stage 2.
In stage 2, we employ multi-scale attention to further extract the detailed information of the degraded images more precisely. 

\subsection*{Real-like Cataract Fundus Image Synthesis Stage}
In order to synthesize cataract fundus images that are closer to real-world scenarios, we propose a real-like cataract fundus image synthesis stage, as shown stage 1 in Fig. \ref{fig:p1}. Inspired by ArcNet \cite{li2022annotation}, we utilize the model proposed in \cite{peli1989restoration} to simulate the degradation of cataract fundus images and make improvements accordingly. The simulate formulation is represented as:
\begin{equation}
\ C({s_C}) = \alpha  \cdot {s_C} * {g_B}\left( {{r_B},{\sigma _B}} \right)
 + \beta  \cdot J * {g_L}\left( {{r_L},{\sigma _L}} \right) \cdot \left( {{L_C} - {s_C}} \right),
\end{equation}
where \(C({s_C})\) represents the simulated cataract fundus image, \({s_C}\) denotes the clear image, c stands for the image's r, g, b channel, \(\alpha\) and \(\beta\) represent the weights of the clear fundus image and the noise from the cataract, \(*\) denotes the convolution operation, \({g_B}\) and \({g_L}\) represent gaussian filters for smoothing the clear image and the cataract panel, respectively, \({g}\left( {{r},{\sigma}} \right)\) denotes a gaussian filter with radius r and spatial constant \(\sigma\), J represents the cataract panel, and \({L_C}\) represents the highest intensity of \({s_C}\). 

However, the degradation that can be simulated by mathematical formulas is limited. Therefore, we further synthesize cataract fundus degraded images that are closer to real-world scenarios through an improved CycleGAN \cite{zhu2017unpaired}. Through two GAN networks, we learn mappings from the sythesized cataract image domain s to the real cataract image domain r, and from real cataract image domain r to the synthesized cataract image domain s. We ensure the similarity between generated images and input image content through GAN loss functions and cycle consistency loss functions. The generative networks are adapted from Johnson et al. \cite{johnson2016perceptual}, which contains two convolutional layers with stride 2, several residual blocks and two convolutional layers with stride $\frac{1}{2}$. For the discriminator network, we use 70 $\times$ 70 PatchGANs \cite{isola2017image,li2016precomputed}. In order to retain as much structural information of the fundus image as possible, inspired by ArcNet \cite{li2022annotation}, based on the Retinex theory \cite{yao2018improved}, we utilize a high-frequency extraction module \(H\left( \cdot \right)\) to extract structural information from the fundus image as the guidance. The high-frequency extraction module \(H\left( \cdot \right)\) can be represented as:
\begin{equation}\ H\left( I \right) = I - I * {g_P}\left( {{r_P},{\sigma _P}} \right),\
\end{equation}
where I represents the fundus image, and \({g_P}\left( {{r_P},{\sigma_P}} \right)\) denotes a gaussian filter with radius \(r_P\) and spatial constant \(\sigma_P\).

\begin{figure*}[!t] % Two column figure (notice the starred environment)
\includegraphics[width=1\linewidth]{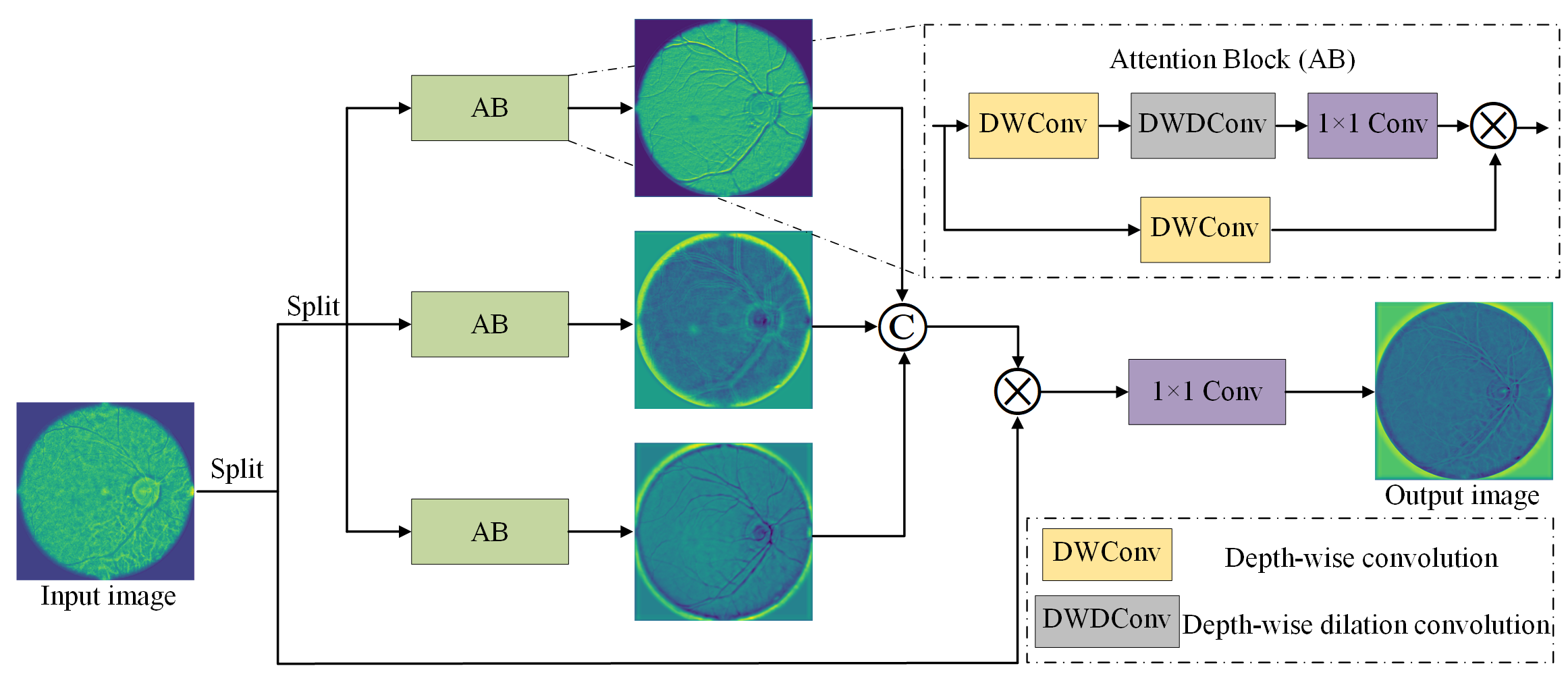}
\caption{The architecture of multi-scale attention block (MSAB). Our MSAB mainly consists of 3 attention blocks (AB) with different scales of convolution kernal to explore the multi-scale features for the better enhancement.}
\label{fig:p2}
\end{figure*}

\subsection*{Multi-scale Attention-Based Cataract Fundus Image Enhancement Stage}

To fully leverage the features of the synthesized real-like cataract fundus images and enhance the enhancement effect, we propose a multi-scale attention-based cataract fundus image enhancement stage, as shown in the stage 2 of Fig. \ref{fig:p1}. This stage consists of a high-frequency extraction module and three multi-scale attention (MSA) modules. The high-frequency extraction module aims to preserve the structural information of the fundus image for better restoration of cataract fundus image details. Inspired by 
 \cite{wang2022multi}, the MSA is designed to exploit the features of the synthesized real-like cataract fundus images and learn richer characteristics of the fundus images. The learning process of the multi-scale attention modules can be represented as follows:
\begin{equation}
    \ {M_i} = {F_i} + {f_{MSAB}}\left( {LN\left( {{F_i}} \right)} \right),\
\end{equation}
\begin{equation}
    \ F_{i + 1} = {M_i} + {f_{GDFB}}\left( {LN\left( {{M_i}} \right)} \right),\
\end{equation}
where \({F_i}\) and \({F_{i+1}}\) represent the input and output features of the multi-scale attention module, \({M_i}\) represents the extracted multi-scale features, \(f_{MSAB}\left( \cdot \right)\) denotes the MSAB module, \(f_{GDFB}\left( \cdot \right)\)  denotes the GDFB module, and \(LN\left( \cdot \right)\)  represents the layer normalization operation.
As shown in Fig. \ref{fig:p2}, the MSAB module consists of three attention blocks (AB). Firstly, the features are divided into two parts along the channel dimension, resulting in features \(F_a\) and \(F_b\). Next, \(F_a\) is further divided into three parts along the channel dimension, yielding \(F_{a_1}\), \(F_{a_2}\), and \(F_{a_3}\), which are then fed into the attention blocks (AB). Each AB module employs different convolutional kernels to extract multi-scale features and capture richer image detail information. Subsequently, the features of different scales are concatenated and multiplied element-wise with \(F_b\) to obtain the multi-scale attention features. AB is shown in Fig. \ref{fig:p2}, which consists of two deep-wise convolutions, a deep-wise dilation convolution, and a 1\(\times\)1 convolution to achieve a larger receptive field for feature extraction. The GDFB firstly adopt a channel-wise splitting to split the input into two halves. We remain one branch while applying a deep-wise convolution operationt to the other branch. Afterwards, we merge the dual cranch together by multiplied to obtain the global spatial information.

\subsection*{Loss Function}
\noindent \textbf{Real-like Cataract Fundus Image Synthesis Stage}
We train the real-like cataract fundus image synthesis stage using the loss function from CycleGAN \cite{zhu2017unpaired}, which can be formulated as follows:
\begin{equation}
\begin{aligned}
{L_{CycleGAN}} =  \lambda {L_{cyc}}\left( {G_S^R,G_R^S} \right)+{L_{GAN}}\left( {G_S^R,{D_R},S,R} \right) +  {L_{GAN}}\left( {G_R^S,{D_S},S,R} \right)  ,
 \end{aligned}
\end{equation}
where S and T respectively represent the source domain and target domain, \(G_{XY}\) represents the mapping function from domain X to Y, \(D_S\) and \(D_T\) represent the discriminators for the source and target domains, \(L_{GAN}\) and \(L_{cyc}\) represent the adversarial loss and cycle consistency loss, \(\lambda\) represents the weight parameter. In addition, we also use the structural loss \(L_R\) in \cite{li2023generic} to preserve the structural information of the fundus. The overall loss function can be expressed as:
\begin{equation}  
\begin{aligned}
L_{total} = \ {L_{CycleGAN}}\left( {\cdot} \right) + {L_R}\left( {S_G}, \mathop{R_G}\limits^\wedge \right) + {L_R}\left( {R_G}, \mathop{S_G}\limits^\wedge \right),
\end{aligned}
\end{equation}
where \(X_G\) represents the structural graph of X.

\noindent \textbf{Multi-scale Attention-based Cataract Fundus Image Enhancement Stage}
To preserve the content information of fundus images, retain their structural details, color brightness, and minimize the occurrence of artifacts during the restoration process, we utilize several loss functions. These include the mean squared error loss \(L_{MSE}\) to constrain structural variations, color loss \(L_{Color}\) to maintain color brightness consistency, total variation loss \(L_{TV}\) and \(L_{1}\) loss to preserve image edges and promote smoothness, and structural similarity index loss \(L_{ssim}\) to recover both brightness and structural details. The overall loss function can be expressed as:
\begin{equation}
\begin{aligned}
{L_{multi}} = {\lambda _{TV}}\Delta (c - y) + {\lambda _{MSE}}||c - y|{|_2} 
 +  {\lambda _{L1}}||c - y|{|_2} 
 +   {\lambda _{ssim}}\mathop {ssim}\limits^ \sim  ((c,y))  +  {\lambda _{Colour}}||\mathop {\max }\limits_{rgb} {c^{rgb}} - \mathop {\max }\limits_{rgb} {y^{rgb}}||,
 \end{aligned}
\end{equation}
where \(\lambda\) represents the hyperparameter, \(rgb\) denotes the r, g, b channels, and \(\mathop {ssim}\limits^ \sim   = 1 - ssim\).

\section*{Experiments}\label{sec4}
\subsection*{Dataset and Evaluation Metrics}
\noindent \textbf{Dataset.}
We train and test our TSMSA-Net on publicly available datasets. Specifically, we use the normal and cataract subsets of the Kaggle dataset to create an unpaired dataset for unsupervised training in stage 1. And in stage 2, we use the normal subset of the Kaggle dataset and the degraded normal subset generated from stage 1 to form a paired dataset for supervised training. All images are resized to 512 $\times$ 512 before being sent to the model. For testing, we use a subset of cataract-labeled images from the ODIR-5K dataset and the cataract subset of the Kaggle dataset.

\noindent \textbf{Evaluation Metrics.}
We assess the effectiveness of our TSMSA-Net by using the natural image quality evaluator (NIQE) \cite{guo2020study} and initial score (IS) \cite{zhao2019data} as metrics to evaluate the image enhancement quality, where lower values of NIQE and higher values of IS indicate better performance.
\subsection*{Implementation Details}
We implement our TSMSA-Net on the PyTorch framework, optimize using the Adam optimizer \cite{kingma2014adam} and train on a single V100 GPU. 
In stage 1, we start with an initial learning rate of 0.0002, and the model is trained for 150 epochs with a linear decay of the learning rate. The batch size is set to 8. 
In stage 2, during the training phase, we start with an initial learning rate of 0.0001, and the model is trained for 100 epochs. The input image size is 512×512, and we random crop size to 256×256 and fed into the network. The batch size is set to 4.
For the loss function, we use the weights as follows: \(\lambda_{TV}\)=1, \(\lambda_{MSE}\)=1, \(\lambda_{Color}\)=0.1, \(\lambda_{ssim}\)=0.1, and \(\lambda_{L_1}\)=0.5.
The MSAB employ convolutional kernels with scales of 7-9-1, 5-7-1, and 3-5-1 for different branches.
During the testing phase, the input image size is set to 512×512, and the batch size is set to 1.

\subsection*{Comparision with State-of-the-art Methods}
\subsubsection*{Quantitative Results}
We compare our TSMSA-Net with six state-of-the-art models, including CycleGAN \cite{zhu2017unpaired}, CofeNet \cite{shen2020modeling}, EnlightenGAN \cite{jiang2021enlightengan}, ArcNet \cite{li2022annotation}, PCENet \cite{liu2022degradation}, and GFENet \cite{li2023generic}. Table \ref{tab:t1} and Table \ref{tab:t2} summarize the comparison results, which are based on the pre-trained models provided by the networks and tested on our dataset. From Table \ref{tab:t1} and Table \ref{tab:t2}, we can clearly observe that our TSMSA-Net achieves the best results in NIQE and IS on Kaggle and ODIR-5K. Concretely, our method surpasses GFENet by 0.15 and 0.09 in NIQE on Kaggle and ODIR-5K respectively, although GFENet is trained on a larger dataset. Although ArcNet \cite{li2022annotation} has seen the test set images during the training process, our method still outperforms ArcNet by 0.87 and 0.15 in NIQE and IS on the Kaggle dataset.
\begin{figure*}[!t] % Two column figure (notice the starred environment)
\includegraphics[width=1\linewidth]{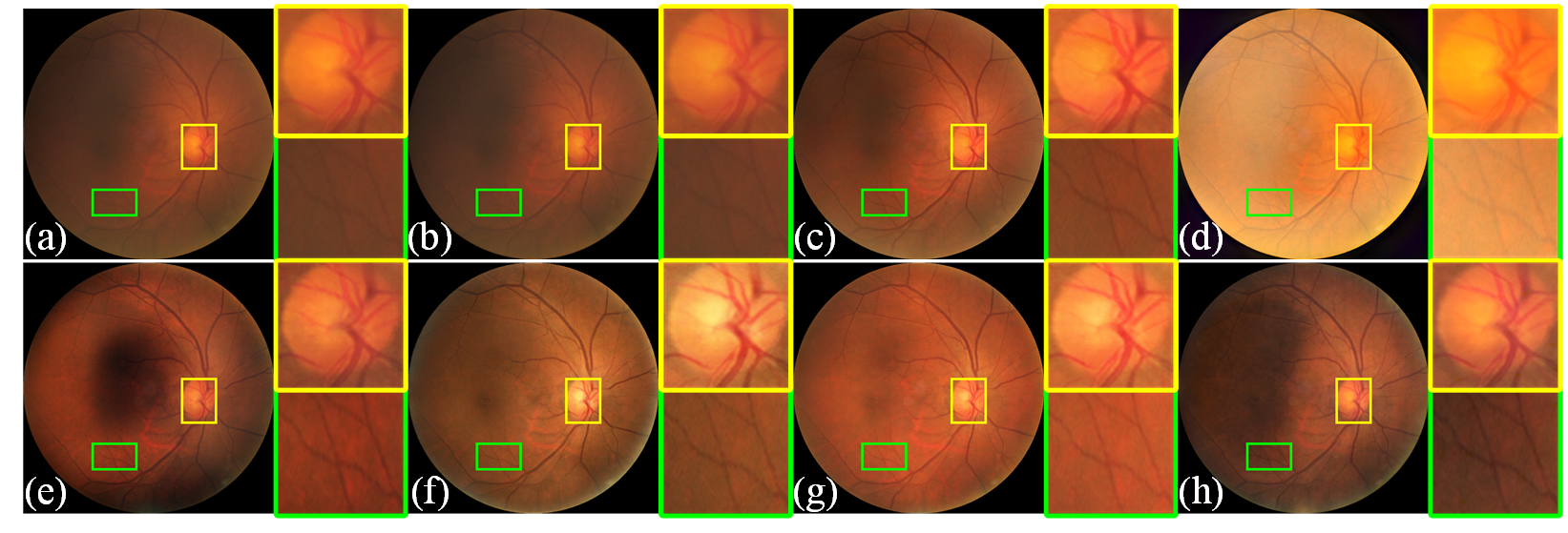}
\caption{Visual comparsion with state-of-the-art methods on Kaggle, and areas of contrast are marked with green and yellow boxes on the original image. (a) cataract image. (b) CycleGAN \cite{zhu2017unpaired}. (c) CofeNet \cite{shen2020modeling}. (d) EnlightenGAN \cite{jiang2021enlightengan}. (e) ArcNet \cite{li2022annotation}. (f) PCENet \cite{liu2022degradation}. (g) GFENet \cite{li2023generic}. (h) TSMSA-Net(Ours).}
\label{fig:p3}
\end{figure*}
\begin{figure*}[!t] % Two column figure (notice the starred environment)
\includegraphics[width=1\linewidth]{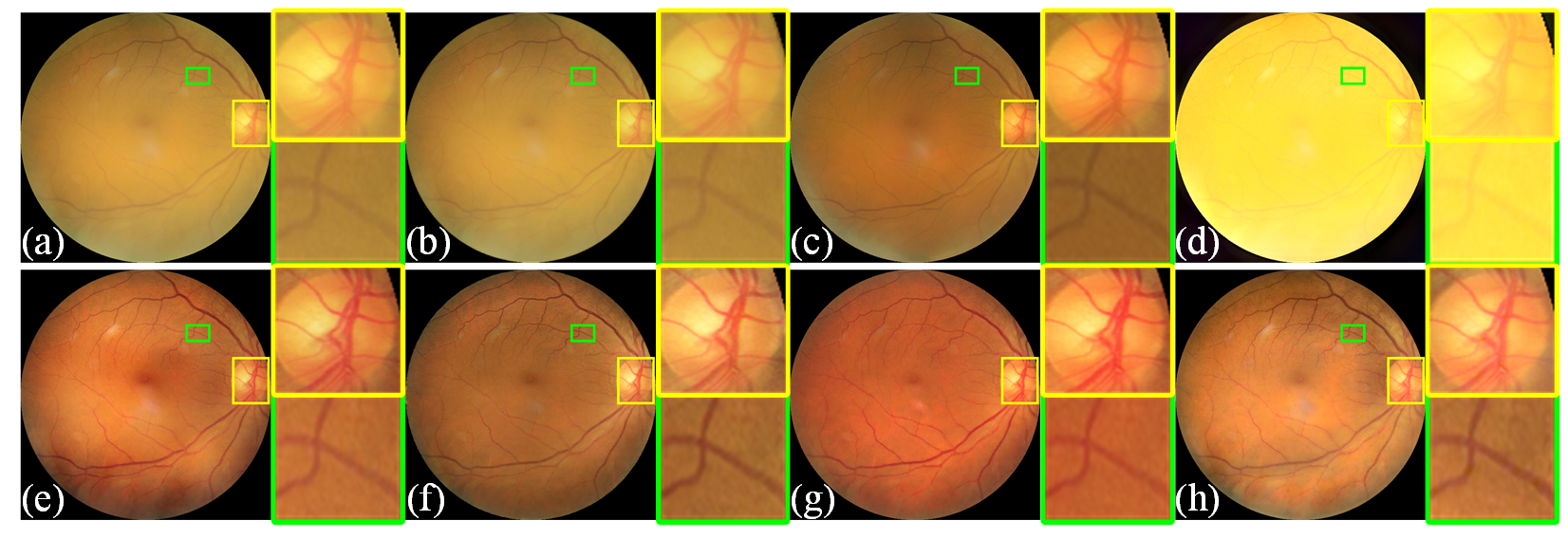}
\caption{Visual comparsion with state-of-the-art methods on ODIR-5K, and areas of contrast are marked with green and yellow boxes on the original image. (a) cataract image. (b) CycleGAN \cite{zhu2017unpaired}. (c) CofeNet \cite{shen2020modeling}. (d) EnlightenGAN \cite{jiang2021enlightengan}. (e) ArcNet \cite{li2022annotation}. (f) PCENet \cite{liu2022degradation}. (g) GFENet \cite{li2023generic}. (h) TSMSA-Net(Ours).}
\label{fig:p4}
\end{figure*}

\begin{table}[t]
\centering
\begin{minipage}{0.48\textwidth}
    \centering
    \begin{tabular}{|l|l|l|}
    \hline
    method & NIQE ↓ & IS ↑ \\
    \hline
    CycleGAN \cite{zhu2017unpaired} & 9.50 & 1.41 \\
    \hline
    CofeNet \cite{shen2020modeling}      & 9.12   & 1.39 \\
    \hline
    EnlightenGAN \cite{jiang2021enlightengan} & 8.95   & 1.49 \\
    \hline
    ArcNet \cite{li2022annotation}       & 7.09   & 1.42 \\
    \hline
    PCENet \cite{liu2022degradation}       & 6.32   & 1.53 \\
    \hline
    GFENet \cite{li2023generic}       & 6.37   & 1.41 \\
    \hline
    TSMSA-Net(Ours)        & \textbf{6.22}   & \textbf{1.57} \\
    \hline
    \end{tabular}
    \caption{Quantitative results on Kaggle dataset. The \textbf{best} results are highlighted in bold.}
    \label{tab:t1}
\end{minipage}
\hfill
\begin{minipage}{0.48\textwidth}
    \centering
    \begin{tabular}{|l|l|l|}
    \hline 
    method           & NIQE ↓ & IS ↑ \\
    \hline 
    CycleGAN \cite{zhu2017unpaired} & 9.85 &1.37 \\
    \hline 
    CofeNet \cite{shen2020modeling}      & 9.75   & 1.39 \\
    \hline 
    EnlightenGAN \cite{jiang2021enlightengan} & 8.47   & 1.50 \\
    \hline 
    ArcNet \cite{li2022annotation}       & 6.43   & 1.31 \\
    \hline 
    PCENet \cite{liu2022degradation}       & 6.20   & 1.53 \\
    \hline 
    GFENet \cite{li2023generic}       & 6.21   & 1.47 \\
    \hline 
    TSMSA-Net(Ours)        & \textbf{6.12}   & \textbf{1.56} \\
    \hline 
    \end{tabular}
    \caption{Quantitative results on ODIR-5K dataset. The \textbf{best} results are highlighted in bold.}
    \label{tab:t2}
\end{minipage}
\end{table}

% \begin{table*}[]
% \centering
% \caption{Quantitative results on Kaggle and ODIR-5K dataset. The \textbf{best} results are highlighted in bold.}
% \begin{tabular}{ccccc}
% \hline
% \multirow{2}{*}{method} & \multicolumn{2}{c}{Kaggle} & \multicolumn{2}{c}{ODIR-5K}\\
% \cline{2-5}& NIQE ↓ & IS ↑& NIQE ↓ & IS ↑ \\
% \hline
% CycleGAN \cite{zhu2017unpaired} & 9.50 & 1.41 & 9.85 &1.37\\
% CofeNet \cite{shen2020modeling}      & 9.12   & 1.39& 9.75   & 1.39  \\
% EnlightenGAN \cite{jiang2021enlightengan} & 8.95   & 1.49 & 8.47   & 1.50 \\
% ArcNet \cite{li2022annotation}       & 7.09   & 1.42 & 6.43   & 1.31 \\
% PCENet \cite{liu2022degradation}       & 6.32   & 1.53  & 6.20   & 1.53\\
% GFENet \cite{li2023generic}       & 6.37   & 1.41   & 6.21   & 1.47 \\
% TSMSA-Net(Ours)        & \textbf{6.22}   & \textbf{1.57} & \textbf{6.12}   & \textbf{1.56}  \\
% \hline
% \label{tab:t1}
% \end{tabular}
% \end{table*}

\subsubsection*{Visual Comparison}
Meanwhile, we also provide the visual comparison between TSMSA-Net and other models on the cataract sub-dataset of Kaggle and the ODIR-5K cataract sub-dataset in Figure \ref{fig:p3} and Figure \ref{fig:p4}, respectively. It is important to note that unlike other models, our model is trained only on the 300 images in the Kaggle sub-dataset, which may contribute to the stylistic differences in the restored images compared to other models. From Figure \ref{fig:p3}, it can be observed that our model is capable of restoring clean and clear vessels in the optic disk, demonstrating the ability of the multi-scale attention module to extract image features. Additionally, as our model has not seen the ODIR-5K dataset during the entire training process, Figure \ref{fig:p4} also reflects the generalization capability of our model. It is evident that our model can effectively restore detailed features such as blood vessels even when trained on a relatively small dataset.

\subsection*{Ablation Study}

In this section, we conduct ablation experiments to investigate the effect of the proposed different components. We test the NIQE \cite{guo2020study} and IS metrics \cite{zhao2019data} on the Kaggle dataset. The experimental results are summarized in Table \ref{tab:t3}. "w/o stage1" indicates that paired images are not obtained through training in  stage 1, and synthetic images generated using the composition function are directly paired with corresponding clear images. "w/o MSAB" means that the multi-scale module is not used in stage 2. We only use a single AB module with a convolution kernel of scale 3-5-1 instead of the three AB modules of different scales.
\begin{table}[t]
\centering
\begin{minipage}{0.48\textwidth}
    \centering
    \begin{tabular}{|l|l|l|l|}
    \hline
    w/o Stage1 & w/o MSAB & NIQE ↓ & IS ↑ \\
    \hline
    \checkmark          &          & 6.28  & 1.53 \\
    \hline
               & \checkmark         & 6.64  & 1.54 \\
    \hline
    \checkmark           & \checkmark         & 6.41  & 1.53 \\
    \hline
               &          & \textbf{6.22}  & \textbf{1.57} \\
    \hline
    \end{tabular}
    \caption{Ablation study on Kaggle dataset. The \textbf{best} results are highlighted in bold.}
    \label{tab:t3}
\end{minipage}
\hfill
\begin{minipage}{0.48\textwidth}
    \centering
    \begin{tabular}{|l|l|l|}
    \hline 
    method           & param (M)&  inference time (s)\\
    \hline 
    CycleGAN \cite{zhu2017unpaired} & 57.1 &11.64 \\
    \hline 
    CofeNet \cite{shen2020modeling}      & 41.2   & 79.72 \\
    \hline 
    EnlightenGAN \cite{jiang2021enlightengan} & 8.6   & 34.84 \\
    \hline 
    ArcNet \cite{li2022annotation}       & 54.4  & 59.29 \\
    \hline 
    PCENet \cite{liu2022degradation}       & 26.6  & 12.63 \\
    \hline 
    GFENet \cite{li2023generic}       & 89.3   & 25.4 \\
    \hline 
    TSMSA-Net(Ours)        & \textbf{0.17}   & \textbf{9.43} \\
    \hline 
    \end{tabular}
    \caption{The comparison of model complexity. The \textbf{best} results are highlighted in bold.}
    \label{tab:t5}
\end{minipage}
\end{table}
\begin{figure}[!t] % Two column figure (notice the starred environment)
\centering
\includegraphics[scale=0.94]{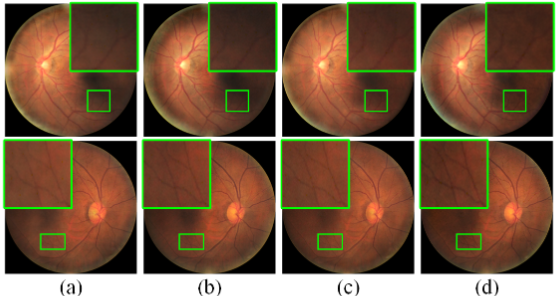}
\caption{Visual comparison on ablation study. (a) represents using synthetic cataract images by simulate formulation directly without the first stage and without multi-scale feature extraction.
(b) represents using synthetic cataract images by simulate formulation directly without the first stage.
(c) represents without multi-scale feature extraction.
(d) represents our method, utilizing both the first stage and multi-scale feature extraction.
The magnified areas are indicated by green boxes in the retinal image.}
\label{fig:p5}
\end{figure}
From the first and fourth rows of Table \ref{tab:t3}, it can be observed that without the learning from stage 1 network, the NIQE and IS indicators decrease by 0.06 and 0.04, respectively. This indicates that the synthesized cataract images from stage 1 are closer to real cataract degradation compared to the synthesized function. As a result, better enhancement effects can be achieved for real cataract images. Moreover, from the second and fourth rows of Table \ref{tab:t3}, it can be seen that employing multi-scale attention significantly boosts performance by 0.42 and 0.03 in NIQE and IS metrics respectively, demonstrating that the extraction of multi-scale information can effectively enhance the ability of image enhancement. Notablely, from the second and third rows of Table \ref{tab:t3}, it can be observed that the effect of using only the image enhanced by the stage 1 network is slightly inferior to directly using the synthesized image. This may be because there is some degree of feature loss during the domain adaptation process. As shown in Table \ref{tab:t3}, using both stage 1 and the multi-scale attention module can achieve the best results on the Kaggle dataset, which can further demonstrate the proposed components's effectiveness.

\begin{figure*}[!h] % Two column figure (notice the starred environment)
\centering
\includegraphics[scale=0.84]{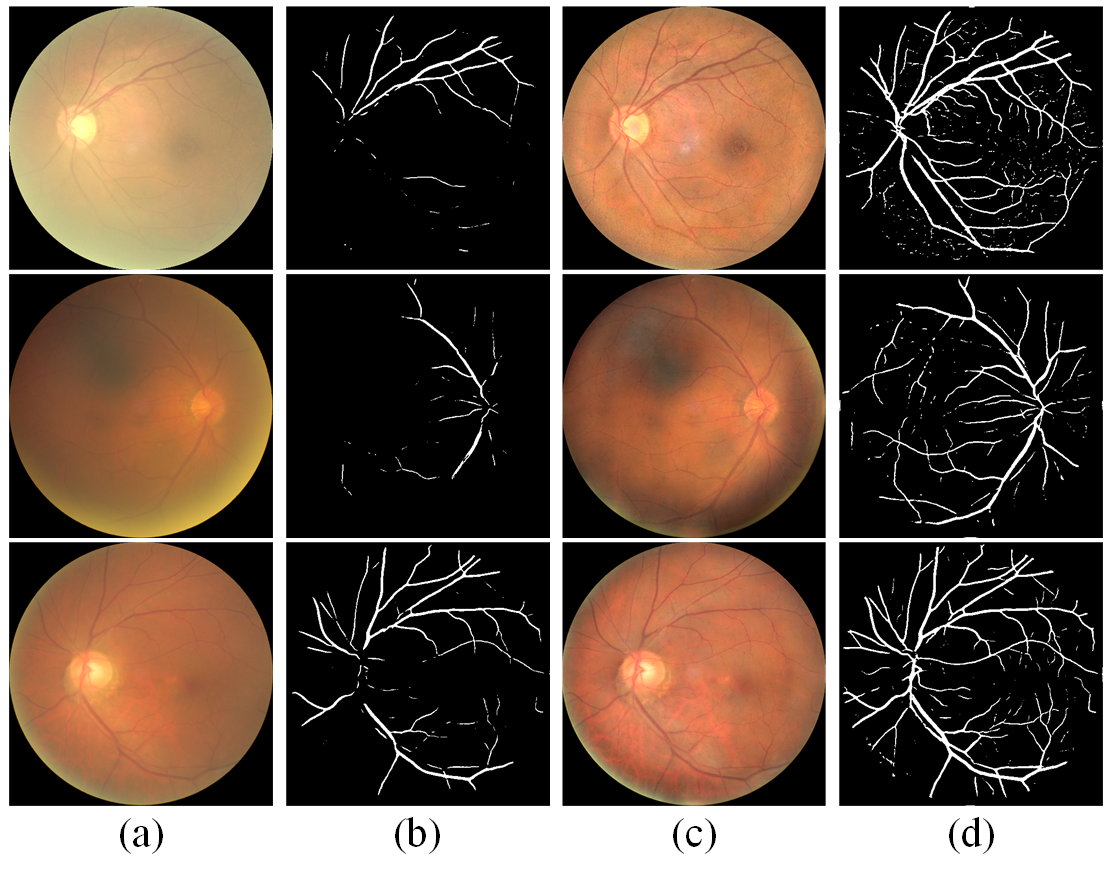}
\caption{Application in vessel segmentation (a) Low-quality image (b) Segmentation result of the low-quality image (c) Enhanced image (d) Segmentation result of the enhanced image}
\label{fig:p7}
\end{figure*}

Furthermore, we also provide visual comparison, as shown in Fig. \ref{fig:p5}. It can be observed that compared with Fig. \ref{fig:p5}(d), without the stage 1 training, Fig. \ref{fig:p5}(a) and Fig. \ref{fig:p5}(b) have more obvious black shadows in the enhanced images. Moreover, comparing the magnified vessel images in Fig. \ref{fig:p5}(c) and Fig. \ref{fig:p5}(d), it shows that the introduction of multi-scale feature extraction can extract richer vessel details, resulting in clearer recovery of vessels.

\subsection*{Model Complexity Comparisons}
An analysis of the model's complexity is essential for a comprehensive evaluation. As shown in Table ~\ref{tab:t5}, a comparative study of parameters and inference time among various models is presented. The inference time is obtained by inferring 100 images on a V100 GPU. In addition, since our model only synthesizes pseudo cataract images in the stage 1, the cataract image enhancement process only occurs in the stage 2, which is equivalent to the image pre-processing process in stage 1. Therefore, when comparing model complexity, only the complexity of the stage 2 of the model is calculated. It can be seen that our method has achieved optimal results in terms of both parameter quantity and inference time. Specifically, the parameter quantity of our model is only 1.9$\%$ of that of GFE-Net, and the inference time is only 37$\%$ of that of GFE-Net. This fully demonstrates the outstanding performance of our method in terms of model complexity.

\begin{table*}[!t]
\centering
\begin{tabular}{|l|l|l|l|l|}
 \hline
\multirow{2}{*}{Disease} & \multicolumn{2}{l|}{Original Image} & \multicolumn{2}{l|}{Enhanced Image} \\
\cline{2-5}
 & Recall & F1 & Recall & F1 \\
 \hline
DR   & 0.61 & 0.73 & 0.65 & 0.77 \\
\hline
ARMD & 0.06 & 0.12 & 0.42 & 0.51 \\
\hline
MH   & 0.79 & 0.81 & 0.94 & 0.96 \\
\hline
BRVO & 0.39 & 0.56 & 0.61 & 0.72 \\
\hline
ODC  & 0.26 & 0.36 & 0.36 & 0.49 \\
\hline
ODE  & 0.29 & 0.43 & 0.76 & 0.72 \\
\hline
\end{tabular}
\caption{Accuracy of automatic disease diagnosis.}
\label{tab:t4}
\vspace{-3mm}
\end{table*}

\subsection*{Applications}
Our enhancement method can be used as the pre-processing for cataract retinal vessel segmentation tasks. We use the U-Net \cite{ronneberger2015u} trained for vessel segmentation on the DRIVE \cite{staal2004ridge} and STARE \cite{hoover2000locating} datasets. The results are shown in Fig. \ref{fig:p7}. It can be observed that before enhancement, the structure of the retina fundus is relatively blurred, making it difficult to effectively segment vessels, resulting in less than ideal segmentation outcomes. However, after our enhancement network, the visibility of vessels in the retinal image is significantly improved, and the segmentation results are noticeably enhanced, effectively improving the ability of retinal vessel segmentation. More vessel structures can be segmented, which is beneficial for further medical diagnosis.

Furthermore, our approach aslo can contribute to automatic disease diagnosis. We use the retinal fundus multi-disease dataset (RFMiD) \cite{pachade2021retinal} to train the ConvNeXt network \cite{liu2022convnet} for automatic detection. RFMiD is created for training automatic classification methods for both common and rare diseases, camprising 3200 fundus images captured by three different fundus cameras. Among them, 317 images are labeled with media haze, which may hinder disease diagnosis. We perform automatic disease detection on the original images and the enhanced images in this chapter. The results are shown in Fig. \ref{fig:p7}. Among them, DR represents diabetic retinopathy, ARMD represents age-related macular degeneration, MH represents media haze, BRVO represents branch retinal vein occlusion, ODC represents optic disc cupping, and ODE represents optic disc edema. We use recall and F1 metrics to evaluate the classification results. The results are shown in Table \ref{tab:t4}. It can be seen that the images enhanced by our model can achieve better classification results than the original images. Specifically, for ARMD, the images enhanced by our method show improvements of 0.36 and 0.39 in recall and F1 metrics, respectively. For ODE, the enhanced images show improvements of 0.47 and 0.29 in recall and F1 metrics, respectively. Especially for images with MH, the images enhanced by our method can achieve recall and F1 metrics of 0.94 and 0.96. This also demonstrates that our enhancement model can contribute to further automated disease detection and diagnosis, while effectively preserving the structural information of the fundus during the enhancement process, thus enhancing the accuracy of automated diagnosis.

\section*{Conclusion}\label{sec5}

In this paper, we have proposed a two-stage multi-scale attention-based network (TSMSA-Net) for weakly supervised cataract fundus image enhancement. To obtain more paired cataract fundus images which are close to the realistic scenarios, we have proposed a real-like cataract fundus image synthesis stage. To better utilize the features of the fundus images, we have proposed a multi-scale attention-based cataract fundus image enhancement stage, which extracts the structural features from different scales to facilitate better image enhancement. Extensive experiments have demonstrated that our TSMSA-Net favors against state-of-the-art cataract fundus image enhancement approaches. Furthermore, TSMSA-Net can improve the results of blood vessel segmentation and automatic disease detection tasks and can improve the accuracy of classification. So it can be used as a pre-processing of computer-aided algorithms for the facilitate diagnosis of ocular diseases. In future work, we will pay more attention to optic disc besides vessels to make the diseases easier for ophthalmologists to distinguish.

\section*{Acknowledgements}
This work was supported in part by the National Key Research and Development Program of China (Grant No. 2021ZD0112400), National Natural Science Foundation of China (Grant No. U1908214), the Program for Innovative Research Team in University of Liaoning Province (Grant No. LT2020015), the Support Plan for Key Field Innovation Team of Dalian (2021RT06), the Support Plan for Leading Innovation Team of Dalian University (XLJ202010), 111 Project (No. D23006), Program for the Liaoning Province Doctoral Research Starting Fund (Grant No. 2022-BS-336), Interdisciplinary project of Dalian University (Grant No. DLUXK-2023-QN-015).

\section*{Author contributions statement}

Xiaoyong Fang. wrote the main manuscript text,  Yue Wang. and Xiangyu Li. conducted the experiments,  Wanshu Fan. and Dongsheng Zhou. analysed the results.  
All authors reviewed the manuscript.

\section*{Data availability}

Te datasets analyzed during the current study are available at ODIR-5K dataset [https://github.com/linhandev/dataset].

\bibliography{sample}

\end{document}